# SVD-Based Restoration of Images Degraded by Shift-Variant Motion Blur

**Arun D. Kulkarni**

**Computer Science Department**

**University of Texas at Tyler, Tyler, TX 75799, USA**

**Email:** akulkarni@uttyler.edu

**Abstract—** Shift-variant image degradation is frequently encountered in practical imaging systems where the point spread function (PSF) varies across the image field due to motion, optical aberrations, atmospheric turbulence, or sensor-related effects. Unlike shift-invariant, shift-variant degradation presents significant challenges for image restoration because the degradation process cannot be represented by a single convolution kernel. This paper proposes a singular value decomposition (SVD)-based framework for restoring images degraded by shift-variant motion blur. The proposed approach determines the contribution of small singular values using a singular-value energy retention criterion. Specifically, the number of small singular values is selected based on a specified percentage of cumulative singular-value energy, providing a systematic approach for controlling noise amplification while preserving useful image information. The degradation model is formulated using a position-dependent PSF represented by a shift-variant imaging operator. Three representative one-dimensional shift-variant motion PSFs are considered: bidirectional linear motion, Gaussian motion, and simple harmonic motion. The image degradation process is modeled as a linear system, and SVD is employed to analyze and invert the corresponding degradation operator. The singular-value representation provides insight into the ill-conditioned nature of the restoration problem and enables the development of stable inversion techniques. The proposed SVD-based restoration algorithm is applied to three degraded images. Experimental results demonstrate the effectiveness of the proposed approach in recovering image details and reducing blur artifacts under different motion models. The proposed framework provides a systematic and computationally effective approach for restoring images affected by shift-variant motion blur and can be extended to more complex degradation scenarios encountered in modern imaging applications.

## 1. Introduction

Image degradation is an inevitable consequence of practical imaging systems and may result from factors such as camera motion, object motion, optical aberrations, atmospheric turbulence, sensor imperfections, and defocusing. The degradation process is commonly characterized by the Point Spread Function (PSF), which describes the response of an imaging system to a point source. In many image restoration problems, the PSF is assumed to be spatially invariant, allowing the degradation process to be modeled as a convolution operation. Although this assumption simplifies the mathematical formulation and restoration process, it is often inadequate for real-world imaging systems in which blur characteristics vary across the image field. In a shift-variant imaging system, the PSF depends on the spatial locations of both the object point and the observation point. Such degradations occur in applications including aerial imaging, astronomical observations, remote sensing, microscopy, medical imaging, and dynamic imaging systems affected by motion. Since the degradation operator varies spatially, conventional restoration techniques based on Fourier-domain deconvolution are not applicable. Instead, the degradation process must be represented by a large linear operator whose structure reflects the position-dependent blur characteristics. Consequently, restoration of shift-variant images becomes a challenging ill-posed inverse problem.

Singular Value Decomposition (SVD) provides a powerful mathematical framework for analyzing and solving ill-conditioned inverse problems. By decomposing the degradation operator into orthogonal singular vectors and corresponding singular values, SVD reveals the fundamental characteristics of the imaging system and provides insight into the loss of image information caused by degradation. Furthermore, regularized SVD-based inversion techniques enable stable image restoration by reducing the influence of small singular values that amplify noise. Shift-variant image restoration is an ill-conditioned inverse problem that is difficult to solve accurately. Numerous approaches based on numerical inversion have been proposed; however, direct inversion often produces unstable solutions because measurement errors, numerical round-off errors, and noise are significantly amplified by the small singular values of the degradation matrix. Although several regularization techniques have been developed for image restoration, methods such as Tikhonov regularization, Wiener filtering, and iterative restoration approaches have certain limitations when applied to shift-variant degradation models. Tikhonov regularization stabilizes the inverse solution by introducing a penalty term that controls the solution norm; however, the selection of the regularization parameter is often problem-dependent and may require additional parameter estimation strategies. Wiener filtering provides an optimal solution under specific assumptions regarding the degradation model and noise statistics, but its effectiveness decreases when the point spread function varies spatially or when accurate noise characteristics are unavailable. Regularization techniques, such as Truncated Singular Value Decomposition (TSVD) and its variants, are commonly employed to overcome this limitation. In the TSVD approach, components associated with small singular values are discarded. However, determining the optimal truncation threshold remains a significant challenge because the restoration quality is extremely sensitive to this parameter. Excessive truncation reduces noise amplification but increases the error between the estimated and original solutions. Conversely, insufficient truncation results in severe noise amplification and degraded restoration performance. Iterative restoration methods, including iterative least-squares and Bayesian approaches, can provide high-quality solutions by incorporating prior information; however, they often involve high computational complexity and require careful initialization and stopping criteria.

In this paper, we propose a modified SVD-based method for shift-variant image restoration. The proposed approach determines the contribution of small singular values using a singular-value energy retention criterion. Specifically, a singular-value energy retention criterion is defined, and the number of retained singular values is selected based on a desired percentage of cumulative singular-value energy. In this work, a 99% energy-retention criterion is employed, providing a systematic and physically meaningful method for selecting the restoration parameters. Unlike conventional TSVD, the proposed method does not eliminate components corresponding to small singular values. Instead, these components are attenuated using weighting factors derived from the singular-value magnitudes, thereby reducing noise amplification while preserving useful image information. The proposed energy-retention-based SVD method directly analyzes the singular-value structure of the degradation operator and provides an explicit representation of the loss caused by the ill-conditioned system. The proposed SVD-based method offers a systematic mechanism for selecting the effective singular components while attenuating noise amplification, making it particularly suitable for shift-variant image restoration problems. To evaluate the proposed method, three shift-variant motion point spread functions (PSFs), bidirectional linear motion, Gaussian motion, and simple harmonic motion (SHM)—are considered. The corresponding degradation matrices are highly ill-conditioned and exhibit different singular-value decay characteristics. An image restoration algorithm is developed and evaluated using these degradation models. Although the same energy-retention criterion is applied in all cases, the number of retained singular values varies, demonstrating the importance of selecting an appropriate threshold for obtaining stable restoration results. The proposed approach is applied to three standard test images, and experimental results are presented and analyzed. The main contributions of this work are twofold: (a) the development of a modified SVD-based image restoration method using a singular-value energy retention criterion, and (b) the formulation of an algorithm for constructing shift-variant motion blur kernels and restoring degraded images using the proposed framework.

The remainder of this paper is organized as follows. Section 2 presents a review of related work. Section 3 describes the mathematical formulation of the shift-variant image degradation model, the motion PSFs considered in this study, and the proposed SVD-based restoration methodology. Section 4 presents experimental results and performance evaluation. Finally, Section 5 summarizes the conclusions and discusses directions for future research.

## 2. Related Work

Image restoration has been extensively studied for recovering images degraded by blur, noise, optical aberrations, and motion effects. Classical restoration techniques are primarily based on inverse filtering, Wiener filtering, and regularized least-squares formulations. Comprehensive treatments of these methods are provided by Gonzalez and Woods [1]. These approaches assume that the degradation process is spatially invariant and can therefore be represented by a convolution operation in the frequency domain. Although such assumptions simplify the restoration process and permit efficient implementation using Fourier transforms, they become inadequate when the point spread function (PSF) varies with spatial position. In many practical imaging systems, including astronomical imaging, remote sensing, microscopy, and dynamic motion imaging, blur characteristics change across the image plane, leading to shift-variant degradation that cannot be accurately modeled by conventional convolution-based techniques.

Restoration of images degraded by spatially varying blur has received considerable attention over the past several decades. One of the early investigations was reported by Guo and Ward [2], who proposed iterative blind restoration methods for images degraded by space-variant blur. Their work demonstrated that image and blur parameters could be estimated simultaneously, even when complete PSF information was unavailable. Nagy and O'Leary [3] subsequently formulated spatially variant image restoration as a large-scale linear inverse problem and developed matrix-based computational techniques for handling non-uniform blur. Their work established the mathematical framework for representing shift-variant degradation operators and remains one of the most influential contributions in this area.

Variational approaches have also been widely used for handling unknown or partially known blurs. Ben Hadj et al. [4] proposed a variational framework for space-variant blind image restoration in which both the latent image and the spatially varying PSF are estimated simultaneously. Zhang et al. [5] introduced a restoration method based on motion-blurred target segmentation, where locally blurred regions are identified and restored independently to improve reconstruction quality. Cheong et al. [6] later developed a computationally efficient framework for spatially varying defocus blur restoration, significantly reducing computational complexity while preserving restoration accuracy.

Image restoration is inherently an ill-conditioned inverse problem because the degradation matrix often possesses singular values that decay rapidly toward zero. Small perturbations caused by measurement noise can therefore result in severe amplification errors in the reconstructed image. Tikhonov and Arsenin [7] introduced regularization theory for stabilizing ill-posed inverse problems through the incorporation of smoothness constraints. Hansen [8-9] provided a rigorous mathematical treatment of rank-deficient and discrete ill-posed problems and established the theoretical basis for truncated singular value decomposition (TSVD), generalized SVD, and Tikhonov regularization. He showed that truncation of small singular values suppresses noise amplification and improves the stability of inverse solutions. His later work further analyzed parameter selection methods and numerical aspects associated with regularized inverse solutions. Several extensions of regularization techniques have been proposed to improve computational efficiency for large-scale problems. Huang et al. [10] combined modified truncated randomized singular value decomposition with Tikhonov regularization to solve high-dimensional inverse problems. Their approach demonstrated improved computational performance while maintaining numerical stability, making it attractive for image restoration applications involving large matrices. Singular value decomposition (SVD) has long been recognized as a powerful mathematical tool for analyzing degradation operators because it reveals the spectral properties and conditioning of inverse problems. Andrews and Hunt [11] emphasized the importance of matrix formulations in image restoration and demonstrated the usefulness of SVD in inverse filtering. SVD provides an orthogonal decomposition of the degradation matrix into singular vectors and singular values, allowing the contribution of different frequency components to be analyzed separately. Rajwade et al. [12] extended matrix decomposition techniques to tensor representations using higher-order singular value decomposition (HOSVD) and demonstrated improved denoising performance while preserving structural image information. These studies indicate that decomposition-based methods provide a physically meaningful and numerically stable framework for solving ill-conditioned restoration problems.

Sparse representation and variational regularization methods have also played a significant role in image restoration. Elad and Aharon [13] introduced the K-SVD algorithm, which learns overcomplete dictionaries for sparse image representation and denoising. Their approach demonstrated that sparse signal models can effectively preserve important image structures while suppressing noise. Rudin, Osher, and Fatemi [14] proposed total variation (TV) regularization, which became one of the most influential edge-preserving restoration techniques. TV-based approaches reduce noise while maintaining discontinuities and image boundaries, making them particularly effective for preserving edges. Numerous optimization-based restoration algorithms developed subsequently have incorporated sparsity constraints, variational formulations, and statistical priors to improve reconstruction quality.

Recent advances in deep learning have revolutionized image restoration. Convolutional neural networks (CNNs) have demonstrated remarkable performance in denoising, deblurring, and super-resolution tasks. Dong et al. [15] proposed SRCNN, one of the earliest CNN architectures for image super-resolution, which established the feasibility of learning restoration mappings directly from data. Zhang et al. [16] introduced DnCNN, which employs residual learning for image denoising and significantly improves restoration accuracy. Ronneberger et al. [17] proposed U-Net, whose encoder-decoder structure and skip connections have inspired numerous restoration networks.

For spatially varying blur, Tian et al. [18] proposed ASANet to restore images affected by wide-field aberrations. Zhang and Zhao [19] introduced decomposition-ascribed synergistic learning for unified image restoration, while Jia et al. [20] incorporated point spread function regularization and active learning to enhance adaptability and restoration performance. More recently, diffusion models have emerged as powerful generative frameworks for image restoration. Li et al. [21] presented a comprehensive review of diffusion-based methods and highlighted their ability to reconstruct fine textures and preserve high-frequency details.

Although considerable progress has been achieved using iterative optimization, sparse representation, and deep learning techniques, several challenges remain in restoring shift-variant images. Many deep-learning approaches require large training datasets and substantial computational resources, while iterative methods often involve complex parameter tuning and high computational cost. Conventional TSVD methods improve numerical stability by discarding small singular values, but the selection of the truncation threshold remains difficult and may result in the loss of useful image information.

The present work investigates a modified SVD-based restoration framework for shift-variant degradation. Instead of eliminating the contribution of small singular values, the proposed method employs an energy-retention criterion and weighting factors derived from singular-value magnitudes to attenuate noise amplification while preserving useful image components. This approach provides a physically interpretable and computationally efficient framework for restoring images degraded by spatially varying motion blur.

## 3. Proposed Framework

The proposed framework addresses the problem of restoring images degraded by a spatially varying or shift-variant point spread function (PSF). Unlike shift-invariant systems, where a single PSF characterizes the entire image, shift-variant degradation involves spatially varying blur characteristics across different image locations. Such degradations commonly occur in imaging systems affected by motion, optical aberrations, atmospheric turbulence, and depth-dependent defocusing. Shift-variant degradation systems result in ill-conditioned inverse problems, and direct inversion often produces unstable or unacceptable solutions due to the amplification of noise and measurement errors. The restoration problem is formulated as an inverse problem and solved using Singular Value Decomposition (SVD). The proposed SVD-based framework provides a robust mechanism for addressing the ill-conditioned nature of image restoration and enables controlled suppression of noise amplification during the inversion process.

### 3.1 Shift Variant Image Degradation Model

The shift variant imaging system is represented by (1).

$$g\left(x',y'\right)=\int_a^b\int_a^b f\left(x,y\right)h\left(x,x';y,y'\right)dydx+\varepsilon\left(x',y'\right) \qquad (1)$$

For $a \le x, y \le b$

Where, $h\left(x,x';y,y'\right)$ represents the shift variant point spread function, $f(x,y)$ represents input image and $\epsilon(x,y)$ is the random additive noise in the in the measurement degraded image $g(x,y)$. Unlike shift-invariant systems, the PSF depends on both the observation point and source point, making the degradation operator spatially varying. For one-dimensional blur, we can consider the blur in x-direction. For certain motion degradation this can be achieved by orienting the co-ordinate system, (1) can be rewritten as

$$g\left(x'\right)=\int_a^b f\left(x\right)h\left(x,x'\right)dx+\varepsilon\left(x'\right) \qquad (2)$$

If the point spread function cannot be reduced to a single dimension, then the technique of stacking can be employed to reduce (1) to (2), which in the form of Fredholm integral Equation of the First Kind. To solve (2) digitally, it can be digitized and written as shown in (3)

$$\mathbf{g} = \mathbf{Hf} + \boldsymbol{\varepsilon} \qquad (3)$$

Where, where $\mathbf{f} \in \mathbb{R}^N$is the original image vector, $\mathbf{g} \in \mathbb{R}^M$is the observed image vector, and $\mathbf{H}$is the shift-variant degradation matrix. and $\boldsymbol{\varepsilon} \in \mathbb{R}^M$ an additive random noise except some restriction on its magnitude (4)

$$\sum \varepsilon_i^2 \le \varepsilon_0^2 \qquad (4)$$

For a discrete image consisting of $N$pixels, each row of $\mathbf{H}$ corresponds to the PSF associated with a particular observation location. The matrix elements are defined as

$$\mathbf{H}_{ij} = h\left(x_j, x_j^{'}\right) \qquad (5)$$

where $x_j$denotes the source pixel location and $x_i'$denotes the observation pixel location. Thus, the degradation matrix takes the form.

$$\mathbf{H} = \begin{bmatrix} h\left(x_1, x_1^{'}\right) & h\left(x_2, x_1^{'}\right) & \dots & h\left(x_N, x_1^{'}\right) \\ h\left(x_1, x_2\right) & h\left(x_1, x_2^{'}\right) & \dots & h\left(x_N, x_2^{'}\right) \\ \dots & \dots & \dots & \dots \\ h\left(x_1, x_M^{'}\right) & h\left(x_2, x_M^{'}\right) & & h\left(x_N, x_M^{'}\right) \end{bmatrix} \qquad (6)$$

Unlike shift-invariant systems, where H is a Toeplitz matrix generated from a single PSF, the shift-variant degradation matrix contains different PSFs in different rows. To construct H, a local PSF is generated for each image location according to the selected blur model (bi-directional linear motion, Gaussian, or SHM blur). The PSF is normalized as below.

$$\sum_{j=1}^{N} h\left(x_i, x_j^{'}\right) = 1, \quad i = 1, 2, \dots M \qquad (7)$$

thereby preserving image energy. The complete degradation matrix is formed by placing the location-dependent PSFs as successive rows, producing a full system matrix that accurately models spatially varying image blur. This matrix representation is particularly suitable for image restoration using Singular Value Decomposition (SVD) since the degradation process can be expressed as a linear inverse problem.

### 3.2 SVD-Based Image Restoration

Equation (3) can be written as

$$\mathbf{f} = \mathbf{H}^{\dagger}(\mathbf{g} - \boldsymbol{\varepsilon}) \qquad (8)$$

Where $\mathbf{H}^{\dagger}$ is the g-inverse of $\mathbf{H}$.

By using singular value decomposition (SVD) of $\mathbf{H}^{\dagger}$ we get

$$\mathbf{f} = \mathbf{U}[\mathbf{D}, \mathbf{0}]\mathbf{V}^{\mathbf{T}}(\mathbf{g} - \boldsymbol{\varepsilon}) \qquad (9)$$

Where columns of $\mathbf{U}$ are eigen vectors of $\mathbf{H}^{\dagger}\mathbf{H}$, the columns of $\mathbf{V}$ are eigen vectors of $\mathbf{H}\mathbf{H}^{\dagger}$ and

$$\mathbf{D} = diag\left[\frac{1}{\omega_1}, \frac{1}{\omega_2}, \ldots, \frac{1}{\omega_N}\right] \qquad (10)$$

Where $\omega_i$ are singular values of matrix $\mathbf{H}$ that are arranged in the descending order that is $\omega_1$ represents the highest singular value and $\omega_N$ represents the smallest singular value**.** From (9) we get

$$f_i = \sum_{j=1}^{N} u_{ij}\beta_j / \omega_j \qquad (11)$$

Where $\boldsymbol{\beta} = \mathbf{V}^{\mathbf{T}}(\mathbf{g} - \boldsymbol{\varepsilon}) \qquad (12)$

In (11), the β-components deviate from their true values due to the presence of noise, **ε**. The solution is extremely sensitive to the β-components associated with the small singular values of the kernel matrix. Shift-variant motion degradation PSFs often produce ill-conditioned kernel matrices, primarily due to the presence of exceedingly small singular values. Consequently, the amplification of noise in the restoration process can be reduced by attenuating the β-components corresponding to the small singular values.

Let $K$ denote the number of small singular values, while $N$ represents the total number of singular values. Therefore, the number of large singular values is ($N$-$K$). The selection of $K$ is critical to the quality of the restored image. If $K$ is chosen too small, the estimated solution approaches the true solution; however, the influence of noise amplification also increases and may eventually render the solution unacceptable. Conversely, if $K$ is chosen too large, important image information may be lost, causing the estimated solution to deviate from the true solution. To determine an appropriate value of $K$, we employ the energy-retention criterion defined in (13). Specifically, $K$ is selected so that 99% of the total singular-value energy is retained.

$$E_R = \left( \sum_{i=1}^{N-K} \omega_i^2 \Big/ \sum_{i=1}^{N} \omega_i^2 \right) \qquad (13)$$

Where $E_R$ represents the fraction of singular-value energy retained. We propose the attenuation factor that is applied to β-components corresponding to small singular values, which reduces the noise amplification due to small singular values. With the above criterion the estimated solution can be defined as

$$\hat{f}_i = \sum_{j=1}^{N-K} u_{ij}\beta_j / \omega_j + \sum_{j=N-K+1}^{N} \phi_i u_{ij}\beta_j / \omega_i \qquad (14)$$

In (14), the β-components associated with the $K$ small singular values are attenuated by the scaling factor $\phi_i = (\omega_i / \omega_K)$. By selecting an appropriate value of $K$, a more accurate estimate of the true solution can be obtained while mitigating the effects of noise amplification in the restored image. We considered three point spread functions for shift variant motion degradation - Bi-directional linear, Gaussian, and simple harmonic motion. These PSFs are described below.

### 3.3 Point Spread Functions for Shift Variant Motion Blur

A bidirectional linear motion blur occurs when the object moves along a straight line during the exposure interval. If the blur length varies with position, the PSF becomes shift-variant. Let L(x)denote the local blur length at position x. The normalized PSF is.

$$h_{BL}\left(x,x^{'}\right)=\begin{cases}\dfrac{1}{L(x)}, & \left|x^{'}-x\right|\le\dfrac{L(x)}{2}\\ 0, & otherwise\end{cases}\qquad(15)$$

The blur length may vary according to

$$L(x)=L_B\left(1+\alpha x\right)\qquad(16)$$

Where $L_0$ is the nominal blur length, and $\alpha$ controls the rate of spatial variation. The normalization condition is.

$$\int_{-\infty}^{+\infty}h_{BL}\left(x,x^{'}\right)dx^{'}=1\qquad(17)$$

Here, $L(x)$may vary spatially due to perspective effects, varying object velocity, or depth-dependent motion. Larger values correspond to stronger blur.

Gaussian motion blur models situations where the probability of displacement follows a Gaussian distribution. The blur spread varies across the image through the position-dependent standard deviation $\sigma(x)$. The PSF is given by.

$$h\left(x,x^{'}\right)=\frac{1}{\left[2\pi\sigma^2(x)\right]^{1/2}}\exp\left(-\frac{\left(x-x^{'}\right)^2}{2\sigma^2(x)}\right)\qquad(18)$$

A commonly used spatially varying standard deviation is

$$\sigma(x) = \sigma_0(1+\gamma x) \tag{19}$$

Where $\sigma_0$ is the nominal blur width, $\gamma$ determines the degree of shift variation. The shape of the blur with shift variant Gaussian PSF is bell shaped that changes with position of the object. Simple harmonic motion (SHM) blur occurs when the imaging sensor or object undergoes oscillatory motion during exposure. Let the instantaneous displacement be.

$$d(t) = A(x)\sin(\omega t) \tag{20}$$

Where $A(x)$ is the position-dependent motion amplitude, $\omega$ is the angular frequency, and $t$ denotes time. The corresponding PSF can be derived from the probability density of the harmonic displacement and is given by.

$$h_{SHM}(x,x') = \begin{cases} \dfrac{1}{\pi\left[A^2(x)-(x'-x)^2\right]^{1/2}}, & |x'-x| < A(x), \\ 0, & |x'-x| \geq A(x) \end{cases} \tag{21}$$

The motion amplitude may vary spatially as

$$A(x) = A_0(1+\gamma x) \tag{22}$$

Where $A_0$ is the nominal oscillation amplitude, and $\gamma$ controls the shift variation.

The proposed approach integrates accurate modeling of spatially varying blur with the numerical robustness of SVD-based inversion, thereby enabling stable image restoration in the presence of ill-conditioned degradation matrices. By attenuating the influence of small singular values, the method effectively reduces noise amplification and yields superior reconstruction quality for shift-variant degradations compared with conventional inverse filtering methods based on spatially invariant PSF assumptions. In our experiments, we considered the three point spread functions described above and generated the corresponding degradation kernels. These kernels were used to degrade three sample images: Lena, Crow, and Text. Random Gaussian noise was subsequently added to the degraded images. The resulting noisy and blurred images were then restored using the modified SVD-based restoration algorithm proposed in the previous section. The restoration results are presented in Section IV.

## 4. Experimental Results

We considered three test images in this experiment: Lena, Crow, and Text Image. The original images are shown in Fig. 1. Each image is size 128 x 128 pixels. Each image was degraded using the shift-variant blur kernels described earlier. To further simulate realistic imaging conditions, random noise corresponding to a maximum signal-to-noise ratio (SNR) of 50.07 dB was added to the degraded images. The presence of noise, combined with the ill-conditioned nature of the degradation operators, makes image restoration by direct inversion highly unstable and often impractical because the corresponding system matrices contain small singular values. The restoration software was developed using MATLAB script to implement the proposed Singular Value Decomposition (SVD)-based approach. The images were degraded using three shift-variant motion-blur Point Spread Functions (PSFs): (a) bi-directional linear blur, (b) Gaussian blur, and (c) Simple Harmonic Motion (SHM) blur. The resulting degraded images and their corresponding restored images obtained using the proposed SVD restoration method are presented in this section.

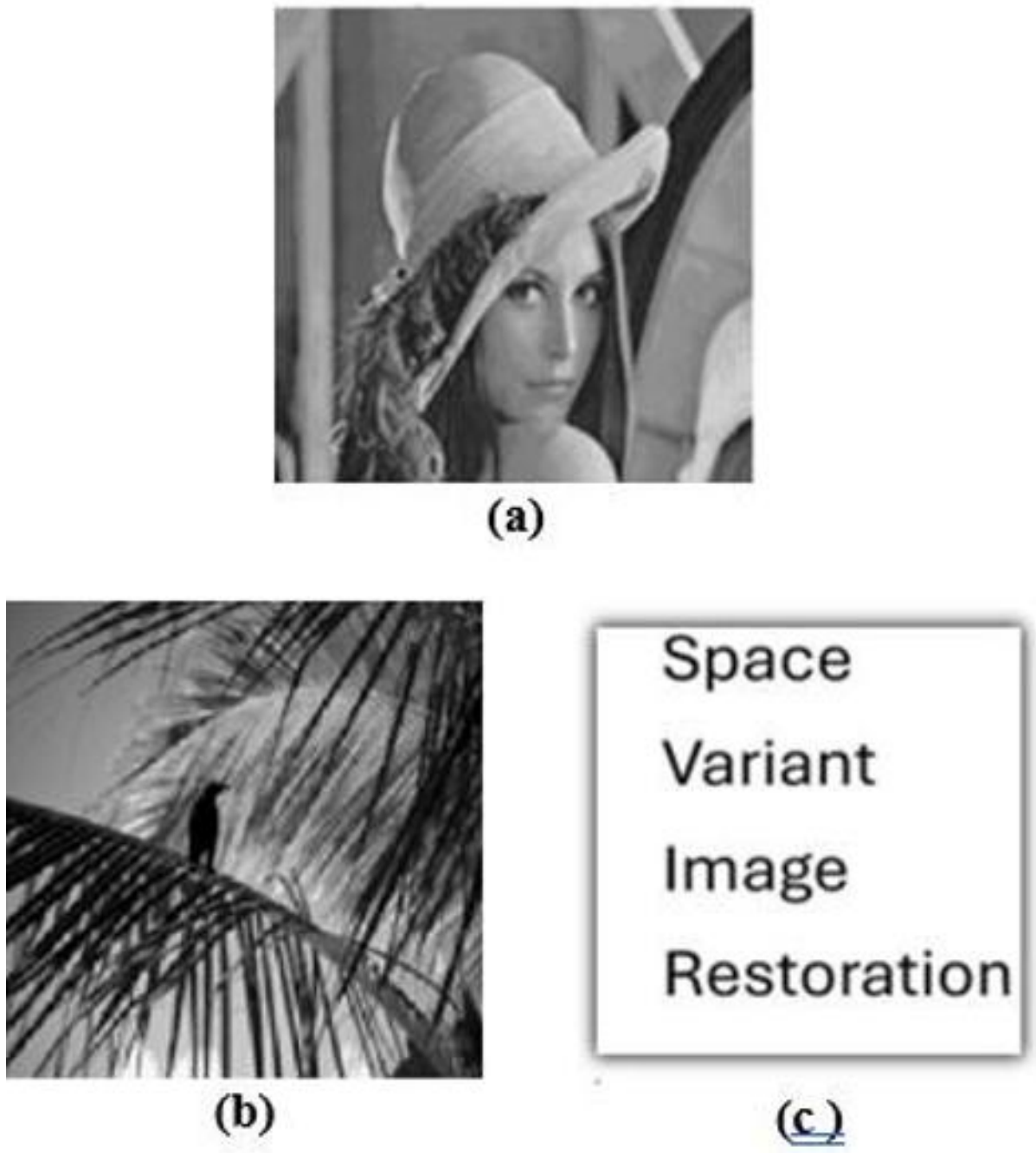

**Fig.1.** Original Images (a) Lena, (b) Crow, (c) Text

## 4.1 Space-Variant Simple Bi-Directional Linear PSF

The kernel matrix obtained from the bi-directional PSF in (15) is shown in Fig. 2. The condition number for bi-directional kernel is 1.2849 x$10^{17}$. The kernel size is 128×128, and the blur length varies from 3 to 21 pixels. In the bi-directional linear motion PSF, the blur length changes with pixel position, resulting in a shift-variant degradation model. The corresponding singular value and singular energy plots are presented in Fig. 3. The total number of singular values is $N$=128.

*1*

In all experiments, a 99% singular energy retention criterion is employed. Under this criterion, the number of small singular values is $K$=64, while the number of large singular values is ($N-K$) =64. In the proposed SVD-based restoration method, the components associated with large singular values are left unchanged, whereas the components corresponding to small singular values are attenuated. The degraded images obtained using bi-directional PSF and their corresponding restored images are shown in Fig. 4.

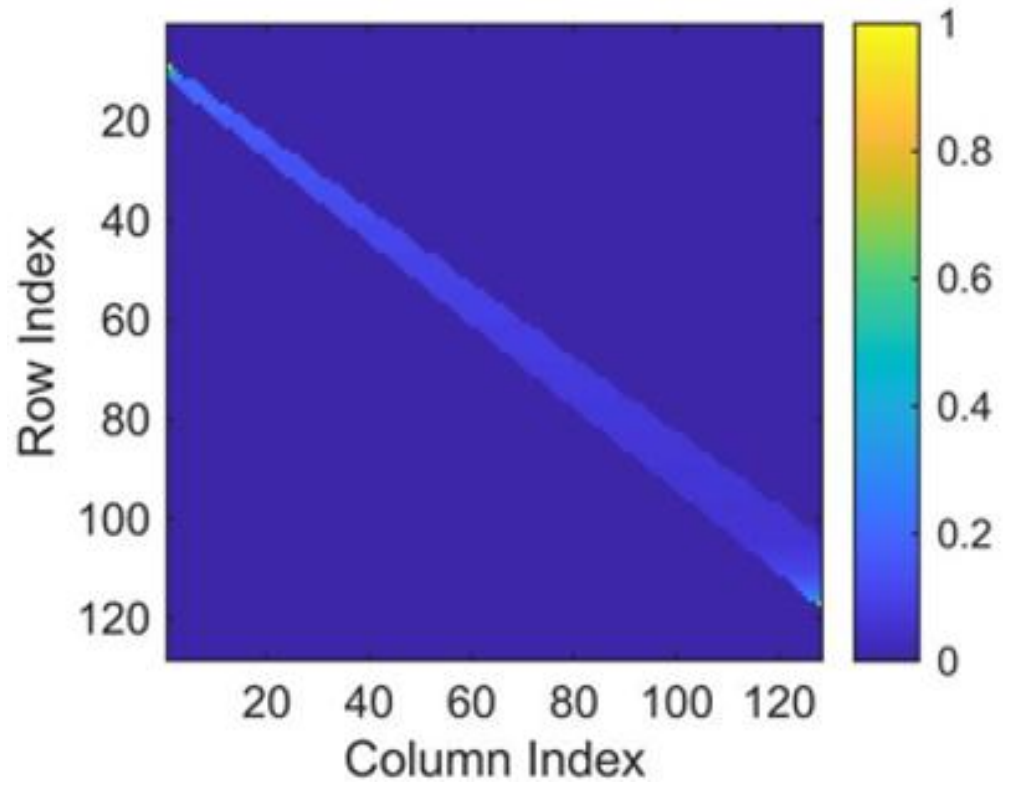

**Fig. 2.** Kernel Matrix for Bi-Directional Linear Blur

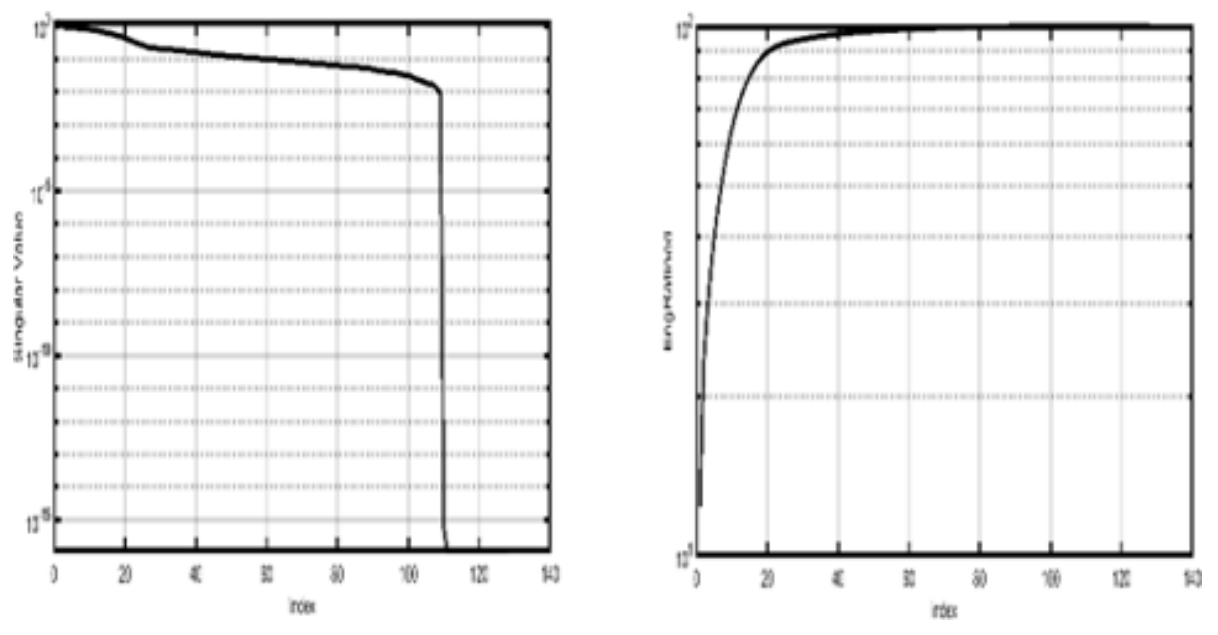

**Fig. 3.** Singular-value and Energy plots for Bi-directional Linear PSF

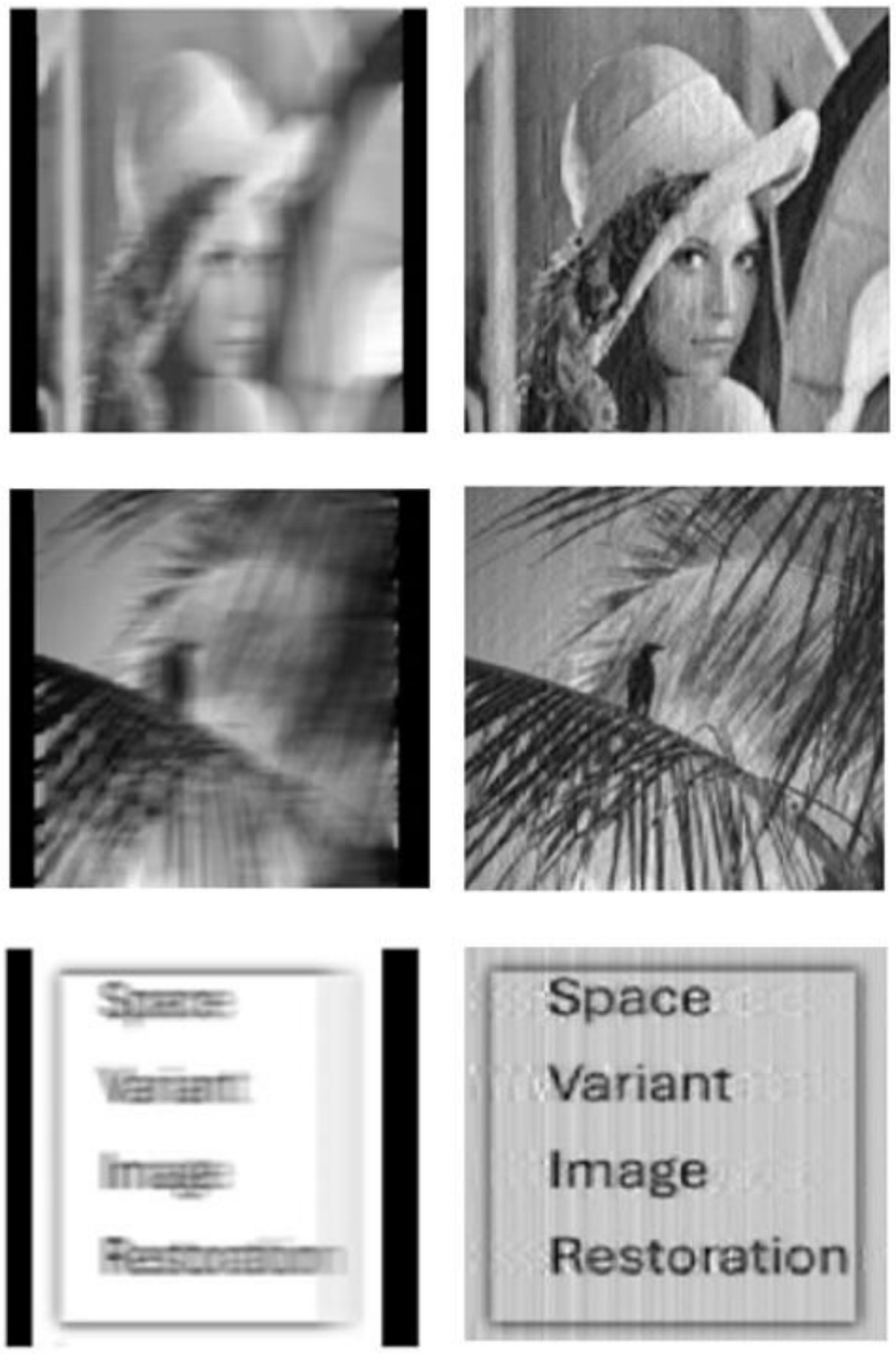

**Fig. 4**. Degraded and Restored Images for Bi-Directional Linear PSF

### 4.2 Space-Variant Gaussian PSF

The kernel matrix obtained from the Gaussian PSF in (18) is shown in Fig. 5. The kernel size is 128×128, and the standard deviation in (18) varies from 1 to 8. In the Gaussian PSF, the variance for the blur changes with pixel position, resulting in a shift-variant degradation model. The corresponding singular value and singular energy plots are presented in Fig. 6. The condition number for Gaussian kernel is 3.442 x$10^8$. The total number of singular values is $N$=128. In all experiments, a 99% singular energy retention criterion is employed. Under this criterion, the number of small singular values is $K$=105, while the number of large singular values is $(N-K)$

=23. In the proposed SVD-based restoration method, the components associated with large singular values are left unchanged, whereas the components corresponding to small singular values are attenuated. The degraded images obtained using the Gaussian PSF and their corresponding restored images are shown in Fig. 7.

### 4.3 Shift-Variant Simple Harmonic Motion PSF

Unlike linear and Gaussian motion PSFs, the SHM PSF exhibits higher weighting near the extreme motion positions, reflecting the longer residence time of the oscillating object at kernel size is 128×128, and the amplitude in (21) varies 2 to 4. In the SHM PSF, the amplitude changes with pixel position, resulting in a shift-variant degradation model. The kernel matrix obtained from the SHM PSF in (21) is shown in Fig. 8. The kernel size is 128×128, and the amplitude in (21) varies from 2 to 4. The corresponding singular value and singular energy plots are presented in Fig. 9. The condition number for SHM kernel is 1.4752 $x10^4$.

The total number of singular values is $N$=128. In all experiments, a 99% singular energy retention criterion is employed. Under this criterion, the number of small singular values is $K$=42, while the number of large singular values is $(N-K)$ =86. In the proposed SVD-based restoration method, the components associated with large singular values are left unchanged, whereas the components corresponding to small singular values are attenuated. The degraded images obtained using the SHM PSF and their corresponding restored images are shown in Fig. 10.

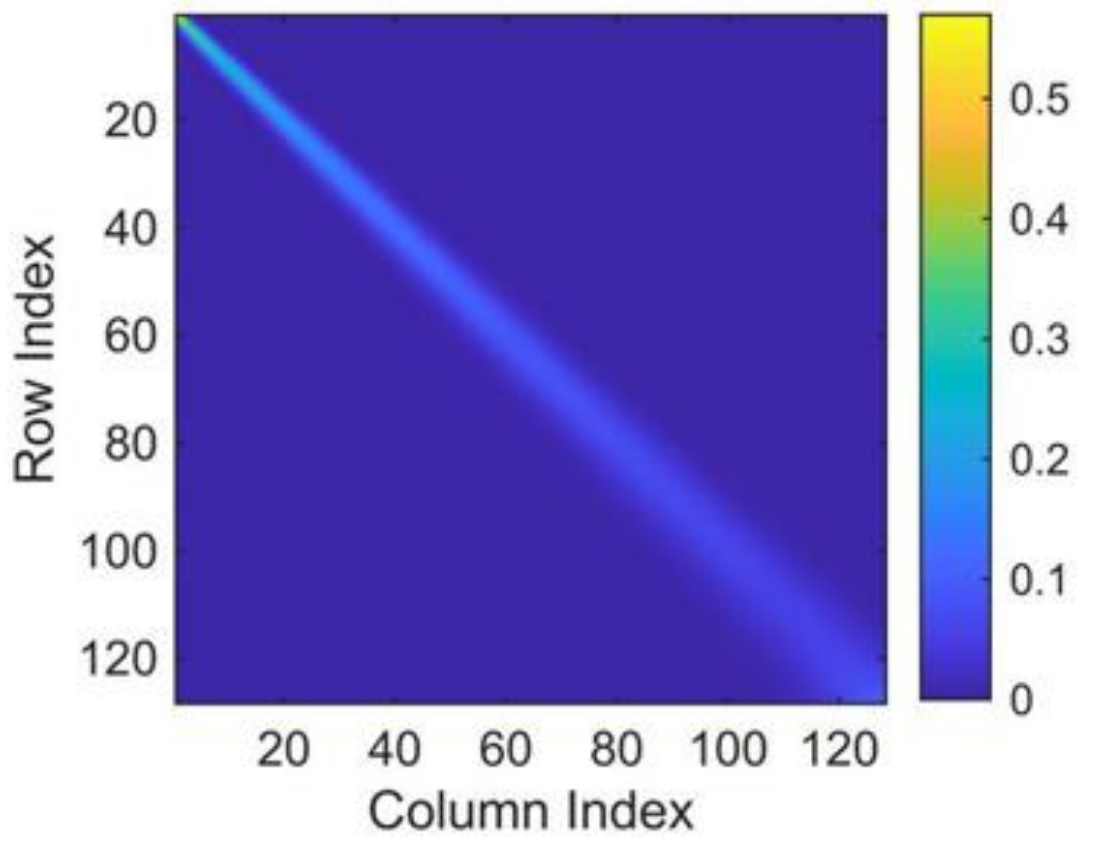

**Fig. 5.** Kernel Matrix for Gaussian Blur

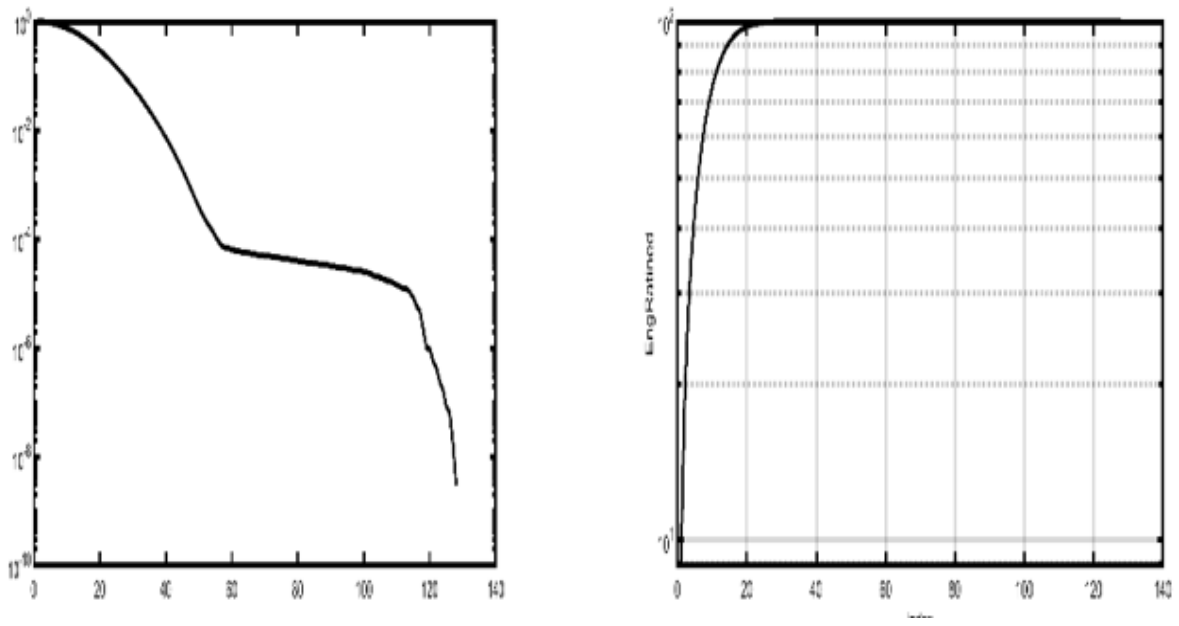

**Fig. 6**. Singular-value and Energy plots for Gaussian PSF

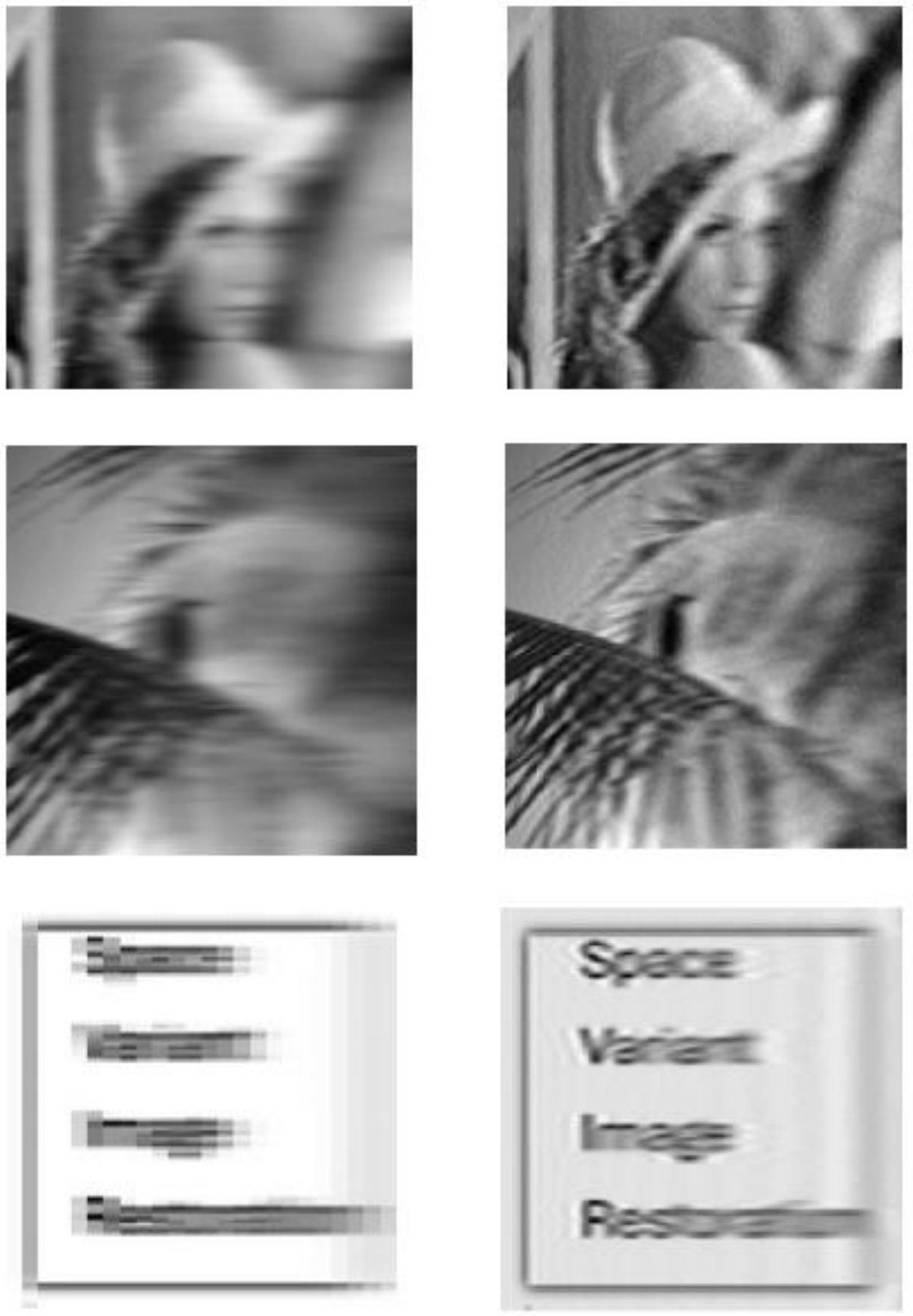

**Fig. 7.** Degraded and Restored Images with Gaussian PSF

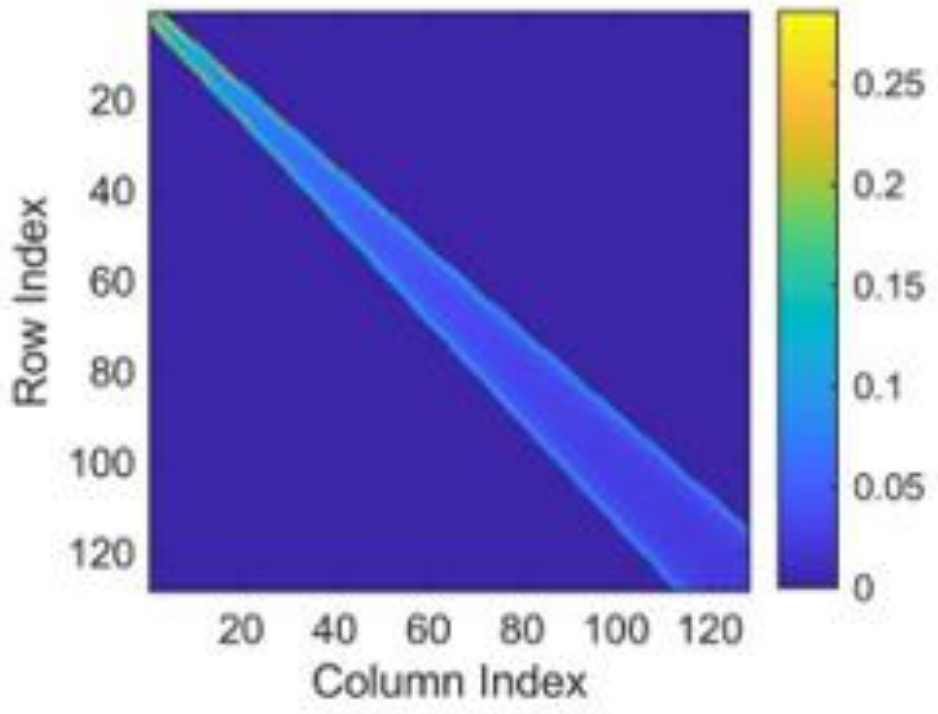

**Fig. 8.** Kernel Matrix for SHM PSF

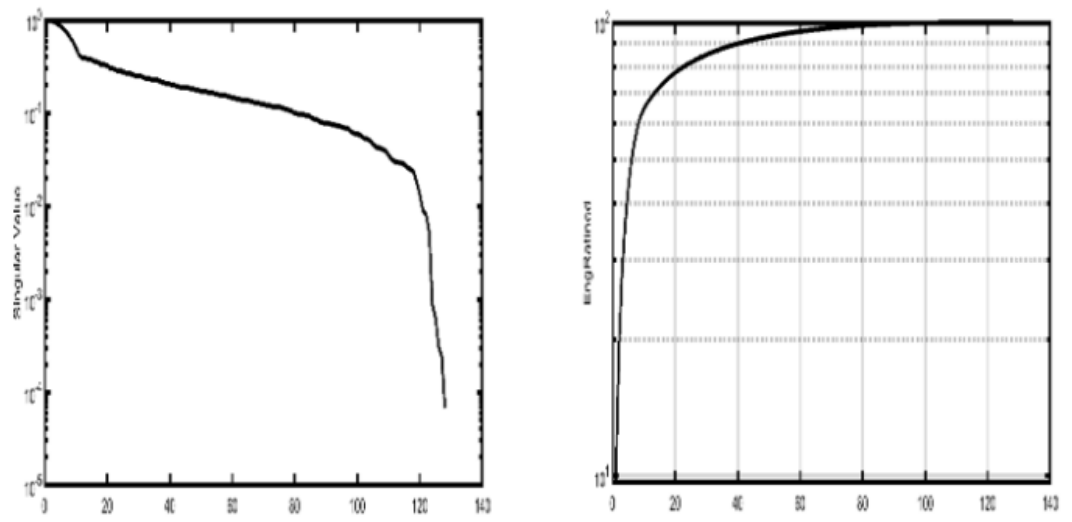

**Fig. 9**. Singular-value and Energy plots for SHM PSF

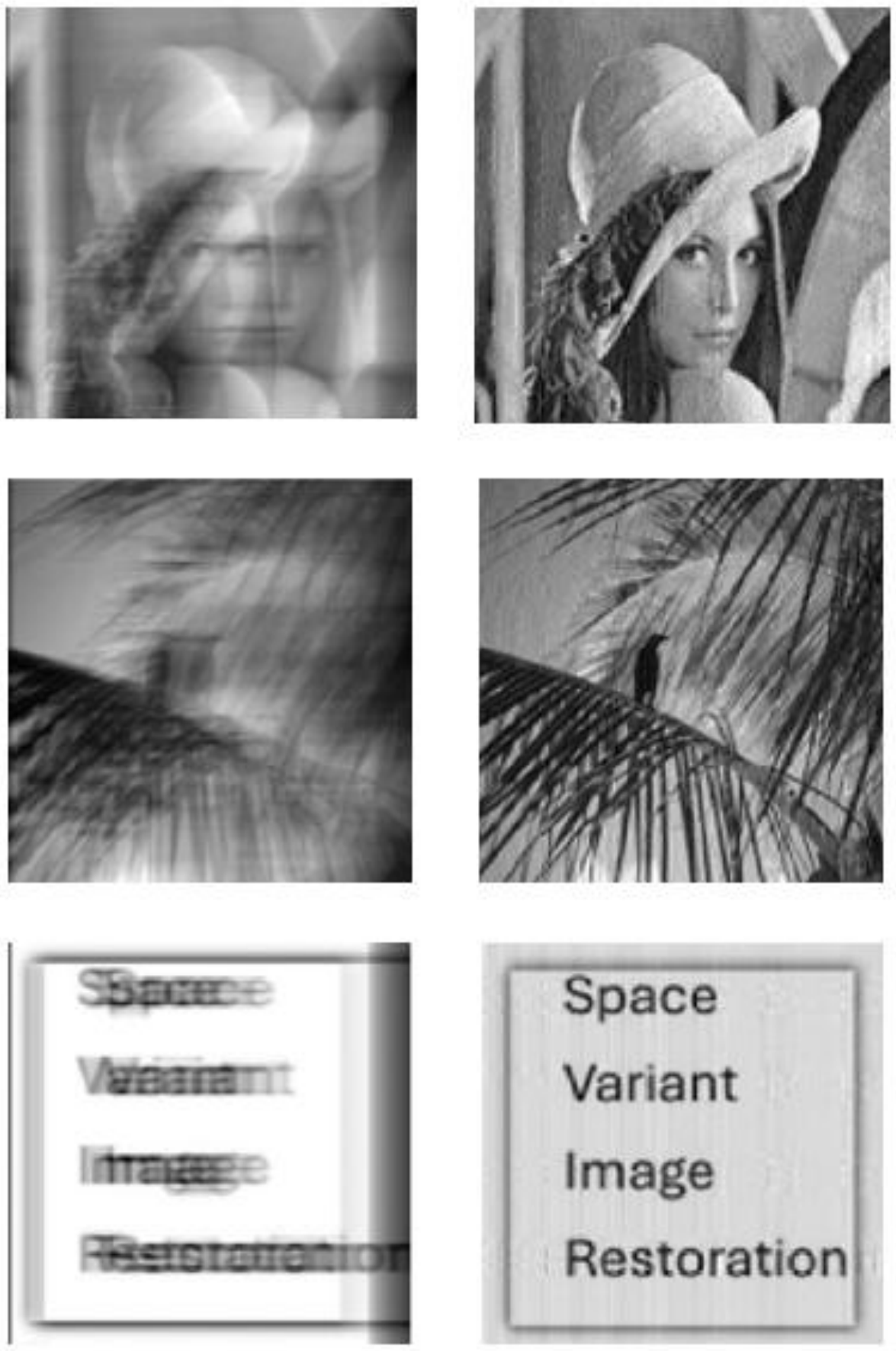

**Fig. 10.** Degraded and Restored Images with SHM PSF

## 5. Conclusion

This paper presented a singular value decomposition (SVD)-based framework for the restoration of images degraded by shift-variant blur. Unlike shift-invariant degradation, shift-variant image degradation cannot be represented by a single convolution operation and therefore leads to large, ill-conditioned inverse problems. To address this difficulty, a modified SVD-based restoration approach was developed in which the contribution of small singular values is controlled through a singular-value energy retention criterion. By selecting the effective singular components according to the cumulative singular-value energy and attenuating the influence of small singular values, the proposed method provides a stable inverse solution while suppressing noise amplification.

Three representative shift-variant motion point spread functions—bidirectional linear motion, Gaussian motion, and simple harmonic motion (SHM)were considered to model different spatially varying blur characteristics. The corresponding degradation operators exhibited distinct singular-value distributions and varying degrees of ill-conditioning. Experimental results obtained using three test images demonstrated that the proposed approach effectively restored blurred images and recovered important structural details under different degradation conditions. The energy-retention-based strategy provided a systematic and physically meaningful mechanism for selecting the effective singular components, thereby improving restoration stability, and avoiding the sensitivity associated with conventional truncation methods.

Overall, the proposed framework combines the analytical advantages of SVD with an energy-based regularization mechanism, providing an effective and computationally efficient approach for shift-variant image restoration. Future work will focus on extending the method to two-dimensional spatially varying point spread functions, incorporating adaptive and data-driven parameter selection strategies, and integrating the proposed framework with iterative and deep-learning-based restoration methods to address more complex degradation scenarios encountered in modern imaging systems.

**Data Availability** No datasets were generated or analyzed during the current study.